\documentclass{article}
\pdfoutput=1
\usepackage{nips14submit_e,times}

\usepackage{graphicx}
\usepackage{subfig} 
\usepackage{natbib}
\usepackage{algorithm}
\usepackage[noend]{algorithmic}
\usepackage{amsmath}
\usepackage{amsthm}
\usepackage{amssymb}
\usepackage{multirow}
\usepackage{hyperref}
\hypersetup{
    colorlinks,
    citecolor=gray,
    linkcolor=blue,
    urlcolor=blue
}

\newcommand{\E}{\mathbb{E}}
\newcommand{\ind}{\boldsymbol{1}}

\renewcommand{\L}{\mathcal{L}}

\renewcommand{\P}{\mathcal{P}}

\newcommand{\W}{\mathcal{W}}

\newcommand{\Y}{\mathcal{Y}}

\newcommand{\bzero}{\boldsymbol{0}}
\newcommand{\bone}{\boldsymbol{1}}

\newcommand{\blambda}{\boldsymbol{\lambda}}

\renewcommand{\v}{\boldsymbol{v}}

\newcommand{\st}{\mathrm{s.t.}}

\newcommand{\diag}{\mathrm{diag}}

\newcommand{\KL}{\mathbf{KL}}

\newcommand{\df}[2]{\frac{\partial #1}{\partial #2}}

\newtheorem{theorem}{Theorem}
\newtheorem{lemma}[theorem]{Lemma}

\newcommand{\seclabel}[1]{\label{sec:#1}}
\newcommand{\secref}[1]{Section~\ref{sec:#1}}
\newcommand{\secsref}[2]{Sections~\ref{sec:#1} and \ref{sec:#2}}

\newcommand{\figlabel}[1]{\label{fig:#1}}
\newcommand{\figref}[1]{Figure~\ref{fig:#1}}

\newcommand{\eqlabel}[1]{\label{eq:#1}}
\renewcommand{\eqref}[1]{Equation~(\ref{eq:#1})}
\newcommand{\eqsref}[2]{Equations~(\ref{eq:#1}) and (\ref{eq:#2})}

\newcommand{\lemlabel}[1]{\label{lem:#1}}
\newcommand{\lemref}[1]{Lemma~\ref{lem:#1}}

\newcommand{\alglabel}[1]{\label{alg:#1}}
\newcommand{\algref}[1]{Algorithm~\ref{alg:#1}}

\newcommand{\applabel}[1]{\label{app:#1}}
\newcommand{\appref}[1]{Appendix~\ref{app:#1}}

\makeatletter
\let\save@mathaccent\mathaccent
\newcommand*\if@single[3]{%
  \setbox0\hbox{${\mathaccent"0362{#1}}^H$}%
  \setbox2\hbox{${\mathaccent"0362{\kern0pt#1}}^H$}%
  \ifdim\ht0=\ht2 #3\else #2\fi
  }
\newcommand*\rel@kern[1]{\kern#1\dimexpr\macc@kerna}
\newcommand*\widebar[1]{\@ifnextchar^{{\wide@bar{#1}{0}}}{\wide@bar{#1}{1}}}
\newcommand*\wide@bar[2]{\if@single{#1}{\wide@bar@{#1}{#2}{1}}{\wide@bar@{#1}{#2}{2}}}
\newcommand*\wide@bar@[3]{%
  \begingroup
  \def\mathaccent##1##2{%
    \let\mathaccent\save@mathaccent
    \if#32 \let\macc@nucleus\first@char \fi
    \setbox\z@\hbox{$\macc@style{\macc@nucleus}_{}$}%
    \setbox\tw@\hbox{$\macc@style{\macc@nucleus}{}_{}$}%
    \dimen@\wd\tw@
    \advance\dimen@-\wd\z@
    \divide\dimen@ 3
    \@tempdima\wd\tw@
    \advance\@tempdima-\scriptspace
    \divide\@tempdima 10
    \advance\dimen@-\@tempdima
    \ifdim\dimen@>\z@ \dimen@0pt\fi
    \rel@kern{0.6}\kern-\dimen@
    \if#31
      \overline{\rel@kern{-0.6}\kern\dimen@\macc@nucleus\rel@kern{0.4}\kern\dimen@}%
      \advance\dimen@0.4\dimexpr\macc@kerna
      \let\final@kern#2%
      \ifdim\dimen@<\z@ \let\final@kern1\fi
      \if\final@kern1 \kern-\dimen@\fi
    \else
      \overline{\rel@kern{-0.6}\kern\dimen@#1}%
    \fi
  }%
  \macc@depth\@ne
  \let\math@bgroup\@empty \let\math@egroup\macc@set@skewchar
  \mathsurround\z@ \frozen@everymath{\mathgroup\macc@group\relax}%
  \macc@set@skewchar\relax
  \let\mathaccentV\macc@nested@a
  \if#31
    \macc@nested@a\relax111{#1}%
  \else
    \def\gobble@till@marker##1\endmarker{}%
    \futurelet\first@char\gobble@till@marker#1\endmarker
    \ifcat\noexpand\first@char A\else
      \def\first@char{}%
    \fi
    \macc@nested@a\relax111{\first@char}%
  \fi
  \endgroup
}
\makeatother

\newcommand{\KA}{KA}
\newcommand{\EM}{EM}

\title{Expectation-Maximization \\ for Learning Determinantal Point Processes}

\author{
Jennifer Gillenwater \\
Computer and Information Science \\
University of Pennsylvania \\
\texttt{jengi@cis.upenn.edu} \\
\And
Alex Kulesza \\
Computer Science and Engineering \\
University of Michigan \\
\texttt{kulesza@umich.edu} \\
\And
Emily Fox \\
Statistics \\
University of Washington \\
\texttt{ebfox@stat.washington.edu} \\
\And
Ben Taskar \\
Computer Science and Engineering \\
University of Washington \\
\texttt{taskar@cs.washington.edu} \\
}

\nipsfinalcopy

\begin{document} 

\maketitle

\begin{abstract}
  A determinantal point process (DPP) is a probabilistic model of set
  diversity compactly parameterized by a positive semi-definite kernel
  matrix.  To fit a DPP to a given task, we would like to learn the
  entries of its kernel matrix by maximizing the log-likelihood of the
  available data.  However, log-likelihood is non-convex in the
  entries of the kernel matrix, and this learning problem is
  conjectured to be NP-hard \cite{kulesza2012thesis}.  Thus, previous
  work has instead focused on more restricted convex learning
  settings: learning only a single weight for each row of the kernel
  matrix \cite{kulesza2011uai}, or learning weights for a linear
  combination of DPPs with fixed kernel matrices
  \cite{kulesza2011icml}.  In this work we propose a novel algorithm
  for learning the full kernel matrix.  By changing the kernel
  parameterization from matrix entries to eigenvalues and
  eigenvectors, and then lower-bounding the likelihood in the manner
  of expectation-maximization algorithms, we obtain an effective
  optimization procedure.  We test our method on a real-world product
  recommendation task, and achieve relative gains of up to 16.5\% in
  test log-likelihood compared to the naive approach of maximizing
  likelihood by projected gradient ascent on the entries of the kernel
  matrix.
\end{abstract}

\section{Introduction}
\seclabel{intro}

Subset selection is a core task in many real-world applications.  For
example, in product recommendation we typically want to choose a small
set of products from a large collection; many other examples of subset
selection tasks turn up in domains like document summarization
\cite{lin2012uai,kulesza2012ftml}, sensor placement
\cite{krause2008jmlr,krause2005uai}, image search
\cite{kulesza2011icml,affandi2014icml}, and auction revenue
maximization \cite{dughmi2009ec}, to name a few.  In these
applications, a good subset is often one whose individual items are
all high-quality, but also all distinct.  For instance, recommended
products should be popular, but they should also be diverse to
increase the chance that a user finds at least one of them
interesting.  Determinantal point processes (DPPs) offer one way to
model this tradeoff; a DPP defines a distribution over all possible
subsets of a ground set, and the mass it assigns to any given set is a
balanced measure of that set's quality and diversity.

Originally discovered as models of fermions
\cite{macchi1975aap}, DPPs have recently been effectively
adapted for a variety of machine learning tasks \cite{affandi2014icml,
  snoek2013nips, kang2013nips, affandi2013nips, shah2013uai,
  affandi2013aistats, gillenwater2012nips, zou2012nips,
  affandi2012uai, gillenwater2012emnlp, kulesza2011uai,
  kulesza2011icml, kulesza2010nips}.  They offer attractive
computational properties, including exact and efficient normalization,
marginalization, conditioning, and sampling
\cite{hough2006ps}.  These properties arise in part from
the fact that a DPP can be compactly parameterized by an $N \times N$
positive semi-definite matrix $L$.  Unfortunately, though, learning
$L$ from example subsets by maximizing likelihood is conjectured to be
NP-hard \cite[Conjecture 4.1]{kulesza2012thesis}.  
While gradient ascent can be applied in an attempt to approximately
optimize the likelihood objective, we show later that it requires a
projection step that often produces degenerate results.

For this reason, in most previous work only partial learning of $L$
has been attempted.  \cite{kulesza2011uai} showed that the problem of
learning a scalar weight for each row of $L$ is a convex optimization
problem.  This amounts to learning what makes an item high-quality,
but does not address the issue of what makes two items
similar.  \cite{kulesza2011icml} explored a different direction,
learning weights for a linear combination of DPPs with fixed $L$s.
This works well in a limited setting, but requires storing a
potentially large set of kernel matrices, and the final distribution
is no longer a DPP, which means that many attractive computational
properties are lost. \cite{affandi2014icml} proposed as an alternative
that one first assume $L$ takes on a particular parametric form, and
then sample from the posterior distribution over kernel parameters
using Bayesian methods.  This overcomes some of the disadvantages
of \cite{kulesza2011icml}'s $L$-ensemble method, but does not allow
for learning an unconstrained, non-parametric $L$.

The learning method we propose in this paper differs from those of
prior work in that it does not assume fixed values or restrictive
parameterizations for $L$, and exploits the eigendecomposition of $L$.
Many properties of a DPP can be simply characterized in terms of the
eigenvalues and eigenvectors of $L$, and working with this
decomposition allows us to develop an expectation-maximization (EM)
style optimization algorithm.  This algorithm negates the need for the
problematic projection step that is required for naive gradient ascent
to maintain positive semi-definiteness of $L$.  As the experiments
show, a projection step can sometimes lead to learning a nearly
diagonal $L$, which fails to model the negative interactions between
items.  These interactions are vital, as they lead to the
diversity-seeking nature of a DPP.  The proposed EM algorithm
overcomes this failing, making it more robust to initialization and
dataset changes.  It is also asymptotically faster than gradient
ascent.

\section{Background}
\seclabel{background}

Formally, a DPP $\P$ on a ground set of items $\Y = \{1,\dots,N\}$ is
a probability measure on $2^\Y$, the set of all subsets of $\Y$.  For
every $Y \subseteq \Y$ we have $\P(Y) \propto \det(L_Y)$, where $L$
is a positive semi-definite (PSD) matrix.  The subscript $L_Y \equiv
\left[L_{ij}\right]_{i,j\in Y}$ denotes the restriction of $L$ to the
entries indexed by elements of $Y$, and we have
$\det(L_{\emptyset}) \equiv 1$.  Notice that the restriction to PSD
matrices ensures that all principal minors of $L$ are non-negative, so that
$\det(L_Y) \geq 0$ as required for a proper probability
distribution.  The normalization constant for the distribution can be
computed explicitly thanks to the fact that $\sum_Y \det(L_Y)
= \det(L+I)$, where $I$ is the $N\times N$ identity matrix.  Intuitively, 
we can think of a diagonal entry $L_{ii}$
as capturing the quality of item $i$, while an off-diagonal entry
$L_{ij}$ measures the similarity between items $i$ and $j$.

An alternative representation of a DPP is given by the \emph{marginal
kernel}: $K = L(L + I)^{-1}$.  The $L$-$K$ relationship can also be
written in terms of their eigendecompositons.  $L$ and $K$ share the
same eigenvectors $\v$, and an eigenvalue $\lambda_i$ of $K$
corresponds to an eigenvalue $\lambda_i / (1 - \lambda_i)$ of $L$:
\begin{equation}
K = \sum_{j = 1}^N \lambda_j \v_j \v_j^{\top}\quad\Leftrightarrow\quad
L = \sum_{j = 1}^N \frac{\lambda_j}{1 - \lambda_j} \v_j \v_j^{\top}\;\;.
\eqlabel{k-l-relation}
\end{equation}
Clearly, if $L$ if PSD then $K$ is as well, and the above equations
also imply that the eigenvalues of $K$ are further restricted to be
$\leq 1$.  $K$ is called the \emph{marginal} kernel because, for any
set $Y \sim \P$ and for every $A \subseteq \Y$:
\begin{equation}
\P(A \subseteq Y) = \det(K_A)\,.
\eqlabel{k-marginal}
\end{equation}
We can also write the exact (non-marginal, normalized) probability of
a set $Y \sim \P$ in terms of $K$:
\begin{equation}
\P(Y) = \frac{\det(L_Y)}{\det(L + I)} = |\det(K - I_{\widebar{Y}})|~,
\eqlabel{K-prob}
\end{equation}
where $I_{\widebar{Y}}$ is the identity matrix with entry $(i, i)$
zeroed for items $i \in Y$ \cite[Equation 3.69]{kulesza2012thesis}.
In what follows we use the $K$-based formula for $\P(Y)$ and learn the
marginal kernel $K$.  This is equivalent to learning
$L$, as \eqref{k-l-relation} can be applied to convert from $K$ to
$L$.

\section{Learning algorithms}
\seclabel{algorithm}

In our learning setting the input consists of $n$ example subsets,
$\{Y_1, \ldots, Y_n\}$, where $Y_i \subseteq \{1, \ldots, N\}$ for all
$i$.  Our goal is to maximize the likelihood of these example sets.
We first describe in \secref{baseline} a naive optimization procedure:
projected gradient ascent on the entries of the marginal matrix $K$,
which will serve as a baseline in our experiments.  We then develop an
EM method: \secref{elem-mix} changes variables from kernel entries to
eigenvalues and eigenvectors (introducing a hidden variable in the
process), \secref{em} applies Jensen's inequality to lower-bound the
objective, and \secsref{e-step}{m-step} outline a
coordinate ascent procedure on this lower bound.

\subsection{Projected gradient ascent}
\seclabel{baseline}

The log-likelihood maximization problem, based on \eqref{K-prob}, is:
\begin{equation}
\max_{K} \sum_{i = 1}^n \log\left(|\det(K - I_{\widebar{Y}_i})|\right)\;\;
\st\; K \succeq 0,\;\; I - K \succeq 0
\end{equation}
where the first constraint ensures that $K$ is PSD and the second puts
an upper limit of $1$ on its eigenvalues.  Let $\L(K)$ represent this
log-likelihood objective.  Its partial derivative with respect to $K$
is easy to compute by applying a standard matrix derivative
rule \cite[Equation 57]{matrixcookbook}:
\begin{equation}
\df{\L(K)}{K} = \sum_{i = 1}^n (K - I_{\widebar{Y}_i})^{-1}\,.
\eqlabel{ka-gradient}
\end{equation}
Thus, projected gradient ascent \cite{levitin1966ussr} is a viable,
simple optimization technique.  \algref{ka} outlines this method,
which we refer to as K-Ascent (\KA{}).  The initial $K$ supplied as
input to the algorithm can be any PSD matrix with eigenvalues $\leq
1$.  The first part of the projection step, $\max(\blambda, 0)$,
chooses the closest (in Frobenius norm) PSD matrix to
$Q$ \cite[Equation 1]{henrion2011oms}.  The second part,
$\min(\blambda, 1)$, caps the eigenvalues at $1$.  (Notice that only
the eigenvalues have to be projected; $K$ remains symmetric after the
gradient step, so its eigenvectors are already guaranteed to be real.)

Unfortunately, the projection can take us to a poor local optima.  To
see this, consider the case where the starting kernel $K$ is a poor
fit to the data.  In this case, a large initial step size $\eta$ will
probably be accepted; even though such a step will likely result in
the truncation of many eigenvalues at $0$, the resulting matrix will
still be an improvement over the poor initial $K$.  However, with many
zero eigenvalues, the new $K$ will be near-diagonal, and,
unfortunately, \eqref{ka-gradient} dictates that if the current $K$ is
diagonal, then its gradient is as well.  Thus, the \KA{} algorithm
cannot easily move to any highly non-diagonal matrix.  It is possible
that employing more complex step-size selection mechanisms could
alleviate this problem, but the EM algorithm we develop in the next
section will negate the need for these entirely.

The EM algorithm we develop also has an advantage in terms of
asymptotic runtime.  The computational complexity of \KA{} is
dominated by the matrix inverses of the $\L$ derivative, each of which
requires $O(N^3)$ operations, and by the eigendecomposition needed for
the projection, also $O(N^3)$.  The overall runtime of \KA{}, assuming
$T_1$ iterations until convergence and an average of $T_2$ iterations
to find a step size, is $O(T_1nN^3 + T_1T_2N^3)$.  As we will show in
the following sections, the overall runtime of the EM algorithm is
$O(T_1nNk^2 + T_1T_2N^3)$, which can be substantially better
than \KA{}'s runtime for $k \ll N$.

\begin{figure*}[tb]
\centering
\begin{minipage}[t]{0.48\linewidth}
\begin{algorithm}[H]
\caption{K-Ascent (\KA{})}
 \alglabel{ka}
   \begin{algorithmic}
     \STATE {\bfseries Input:} $K$, $\{Y_1, \ldots, Y_n\}$, $c$
     \REPEAT
       \STATE $G \leftarrow \df{\L(K)}{K}$ (Eq.\ \ref{eq:ka-gradient})
       \STATE $\eta \leftarrow 1$
       \REPEAT
         \STATE $Q \leftarrow K + \eta G$
         \STATE Eigendecompose $Q$ into $V, \blambda$
         \STATE $\blambda \leftarrow \min(\max(\blambda, 0), 1)$
         \STATE $Q \leftarrow V \diag(\blambda) V^{\top}$
         \STATE $\eta \leftarrow \frac{\eta}{2}$
       \UNTIL{$\L(Q) > \L(K)$}
       \STATE $\delta \leftarrow \L(Q) - \L(K)$
       \STATE $K \leftarrow Q$
     \UNTIL{$\delta < c$}
    \STATE {\bfseries Output:} $K$
   \end{algorithmic}
\end{algorithm}
\end{minipage}
\hspace{0.25cm}
\begin{minipage}[t]{0.47\linewidth}
%\vspace{0.25cm}
\begin{algorithm}[H]
\caption{Expectation-Maximization (\EM{})}
 \alglabel{em}
   \begin{algorithmic}
     \STATE {\bfseries Input:} $K$, $\{Y_1, \ldots, Y_n\}$, $c$
     \STATE Eigendecompose $K$ into $V, \blambda$
     \REPEAT
      \FOR{$j = 1, \ldots, N$}
        \STATE $\blambda_j' \leftarrow \frac{1}{n} \sum_{i} p_K(j \in J \mid Y_i)$ (Eq.\ \ref{eq:s-vals})
      \ENDFOR
      \STATE $G \leftarrow \df{F(V, \blambda')}{V}$ (Eq.\ \ref{eq:v-final-em})
      \STATE $\eta \leftarrow 1$
       \REPEAT
         \STATE $V' \leftarrow V \exp[\eta \left(V^{\top}G -G^{\top}V\right)]$
         \STATE $\eta \leftarrow \frac{\eta}{2}$
       \UNTIL{$\L(V', \blambda') > \L(V, \blambda')$}
       \STATE $\delta \leftarrow F(V', \blambda') - F(V, \blambda)$
       \STATE $\blambda \leftarrow \blambda',\;\;V \leftarrow V',\;\;\eta \leftarrow 2\eta$
     \UNTIL{$\delta < c$}
    \STATE {\bfseries Output:} $K$
   \end{algorithmic}
\end{algorithm}
\end{minipage}
\end{figure*}

\subsection{Eigendecomposing}
\seclabel{elem-mix}

Eigendecomposition is key to many core DPP algorithms such as sampling
and marginalization.  This is because the eigendecomposition provides
an alternative view of the DPP as a generative process, which often
leads to more efficient algorithms.  Specifically, sampling a set $Y$
can be broken down into a two-step process, the first of which
involves generating a hidden variable $J \subseteq \{1, \ldots, N\}$
that codes for a particular set of $K$'s eigenvectors.  We review this
process below, then exploit it to develop an EM optimization scheme.

Suppose $K = V \Lambda V^{\top}$ is an eigendecomposition of $K$.  Let
$V^J$ denote the submatrix of $V$ containing only the columns
corresponding to the indices in a set $J \subseteq \{1, \ldots, N\}$.
Consider the corresponding marginal kernel, with all selected
eigenvalues set to $1$:
\begin{equation}
K^{V^J} = \sum_{j \in J} \v_j \v_j^{\top} = V^J (V^J)^{\top}\,.
\eqlabel{elem-kern}
\end{equation}
Any such kernel whose eigenvalues are all $1$ is
called an \emph{elementary} DPP.  According to \cite[Theorem
7]{hough2006ps}, a DPP with marginal kernel $K$ is a mixture of all
$2^N$ possible elementary DPPs:
\begin{equation}
 \P(Y) = \sum_{J \subseteq \{1, \ldots, N\}} \P^{V^J}(Y)
   \prod_{j \in J} \lambda_j \prod_{j \notin J} (1 - \lambda_j)\,,\qquad
 \P^{V^J}(Y) = \ind(|Y| = |J|) \det(K^{V^J}_Y)\,.
   \eqlabel{elem-mix}
\end{equation}
This perspective leads to an efficient DPP sampling algorithm, where a
set $J$ is first chosen according to its mixture weight
in \eqref{elem-mix}, and then a simple algorithm is used to sample
from $P^{V^J}$ \cite[Algorithm 1]{kulesza2012ftml}.  In this sense,
the index set $J$ is an intermediate hidden variable in the process
for generating a sample $Y$.

We can exploit this hidden variable $J$ to develop an EM algorithm for
learning $K$.  Re-writing the data log-likelihood to make the hidden
variable explicit:
\begin{gather}
\L(K) = \L(\Lambda, V) = \sum_{i = 1}^n \log\left(\sum_{J} p_K(J, Y_i)\right) = \sum_{i = 1}^n \log\left(\sum_{J} p_K(Y_i \mid J) p_K(J)\right)\,,\ \ \mathrm{where}
\end{gather}
\begin{align}
p_K(J) =& \prod_{j \in J} \lambda_j \prod_{j \notin J} (1 - \lambda_j)\,,&
p_K(Y_i \mid J) =& \ind(|Y_i| = |J|) \det([V^J (V^J)^\top]_{Y_i})\,.
\eqlabel{v-representation}
\end{align}
These equations follow directly from \eqsref{elem-kern}{elem-mix}.

\subsection{Lower bounding the objective}
\seclabel{em}

We now introduce an auxiliary distribution, $q(J \mid Y_i)$, and
deploy it with Jensen's inequality to lower-bound the likelihood
objective.  This is a standard technique for developing EM schemes for
dealing with hidden variables \cite{neal1998lgm}.  Proceeding in this
direction:
\begin{align}
\L(V, \Lambda) =& \sum_{i = 1}^n \log\left(\sum_{J} q(J \mid Y_i)
\frac{p_K(J, Y_i)}{q(J \mid Y_i)}\right) \geq \sum_{i = 1}^n \sum_{J} q(J \mid Y_i)
\log\left(\frac{p_K(J, Y_i)}{q(J \mid Y_i)}\right) \equiv F(q, V, \Lambda)\,.
\end{align}

The function $F(q, V, \Lambda)$ can be expressed in either of the
following two forms:
\begin{align}
F(q, V, \Lambda) =& \sum_{i = 1}^n -\KL(q(J \mid Y_i) \parallel p_K(J \mid Y_i)) + \L(V, \Lambda) \eqlabel{f1} \\
=& \sum_{i = 1}^n {\E}_q[\log p_K(J, Y_i)] + H(q) \eqlabel{f2}
\end{align}
where $H$ is entropy.  Consider optimizing this new objective by
coordinate ascent.  From \eqref{f1} it is clear that, holding
$V,\Lambda$ constant, $F$ is concave in $q$.  This follows from the
concavity of $\KL$ divergence.  Holding $q$ constant in \eqref{f2}
yields the following function:
\begin{equation}
F(V, \Lambda) = \sum_{i = 1}^n \sum_{J} q(J \mid Y_i) \left[
\log p_K(J) + \log p_K(Y_i \mid J) \right]\,.
\eqlabel{mstep-objective}
\end{equation}
This expression is concave in $\lambda_j$, since $\log$ is concave.
However, it is not concave in $V$ due to the non-convex $V^{\top}V =
I$ constraint.  We describe in \secref{m-step} one way to handle this.

To summarize, coordinate ascent on $F(q, V, \Lambda)$ alternates the
following ``expectation'' and ``maximization'' steps; the first is
concave in $q$, and the second is concave in the eigenvalues:
\begin{align}
\textrm{E-step:}\;\;& \min_{q} \sum_{i=1}^n \KL(q(J \mid Y_i) \parallel p_K(J \mid Y_i)) \\
\textrm{M-step:}\;\;& \max_{V,\Lambda} \sum_{i=1}^n {\E}_q[\log p_K(J, Y_i)]\;\;\st\;\; \bzero \leq \blambda \leq \bone, V^{\top}V = I
\end{align}

\subsection{E-step}
\seclabel{e-step}

The E-step is easily solved by setting $q(J \mid Y_i) = p_K(J \mid
Y_i)$, which minimizes the KL divergence.  Interestingly, we can show
that this distribution is itself a conditional DPP, and hence can be
compactly described by an $N \times N$ kernel matrix.  Thus, to
complete the E-step, we simply need to construct this kernel.
\lemref{q-is-kdpp} (see \appref{lemma1-proof} for a proof) gives an
explicit formula.  Note that $q$'s probability mass is restricted to
sets of a particular size $k$, and hence we call it a $k$-DPP.  A
$k$-DPP is a variant of DPP that can also be efficiently sampled from
and marginalized, via modifications of the standard DPP algorithms.
(See \appref{lemma1-proof} and \cite{kulesza2011icml} for more on
$k$-DPPs.)

\begin{lemma}
At the completion of the E-step, $q(J \mid Y_i)$ with $|Y_i| = k$ is a
$k$-DPP with (non-marginal) kernel $Q^{Y_i}$:
\begin{equation}
Q^{Y_i} = R Z^{Y_i} R,\;\;\mathrm{and}\;\;
q(J \mid Y_i) \propto \ind(|Y_i| = |J|) \det(Q^{Y_i}_J)\,,\ \mathrm{where}
\eqlabel{q-kernel}
\end{equation}
\begin{equation}
U = V^{\top},\;\;\;\;
Z^{Y_i} = U^{Y_i} (U^{Y_i})^\top,\;\;\mathrm{and}\;\;
R = \diag\left(\sqrt{\blambda / (\bone - \blambda)}\right)\,.
\eqlabel{R-val}
\end{equation}
\lemlabel{q-is-kdpp}
\end{lemma}

\subsection{M-step}
\seclabel{m-step}

The M-step update for the eigenvalues is a closed-form expression with
no need for projection.  Taking the derivative
of \eqref{mstep-objective} with respect to $\lambda_j$, setting it
equal to zero, and solving for $\lambda_j$:
\begin{equation}
\lambda_j = \frac{1}{n}\sum_{i = 1}^n \sum_{J : j \in J} q(J \mid Y_i)\,.
\eqlabel{lam-Jexp-up}
\end{equation}
The exponential-sized sum here is impractical, but we can eliminate
it.  Recall from \lemref{q-is-kdpp} that $q(J \mid Y_i)$ is a $k$-DPP
with kernel $Q^{Y_i}$.  Thus, we can use $k$-DPP marginalization
algorithms to efficiently compute the sum over $J$.  More concretely,
let $\hat{V}$ represent the eigenvectors of $Q^{Y_i}$, with
$\hat{v}_r(j)$ indicating the $j$th element of the $r$th eigenvector.
Then the marginals are:
\begin{equation}
\sum_{J : j \in J} q(J \mid Y_i) =
q(j \in J \mid Y_i) = \sum_{r = 1}^N \hat{v}_r(j)^2\,,
\eqlabel{s-vals}
\end{equation}
which allows us to compute the eigenvalue updates in time $O(nNk^2)$,
for $k = \max_i |Y_i|$.  (See \appref{eigval-updates} for the
derivation of \eqref{s-vals} and its computational complexity.)  Note
that this update is self-normalizing, so explicit enforcement of the
$0 \leq \lambda_j \leq 1$ constraint is unnecessary.  There is one
small caveat: the $Q^{Y_i}$ matrix will be infinite if any $\lambda_j$
is exactly equal to 1 (due to $R$ in \eqref{R-val}).  In practice, we
simply tighten the constraint on $\blambda$ to keep it slightly below
1.
% This truncation is technically a projection, but a very small one.

Turning now to the M-step update for the eigenvectors, the derivative
of \eqref{mstep-objective} with respect to $V$ involves an
exponential-size sum over $J$ similar to that of the eigenvalue
derivative.  However, the terms of the sum in this case depend on $V$
as well as on $q(J \mid Y_i)$, making it hard to simplify.  Yet, for the
particular case of the initial gradient, where we have $q = p$,
simplification is possible:
\begin{equation}
\df{F(V, \Lambda)}{V} = \sum_{i=1}^n 2 B_{Y_i} (H^{Y_i})^{-1} V_{Y_i} R^2
\eqlabel{v-final-em}
\end{equation}
where $H^{Y_i}$ is the $|Y_i| \times |Y_i|$ matrix $V_{Y_i} R^2
V_{Y_i}^{\top}$ and $V_{Y_i} = (U^{Y_i})^{\top}$.  $B_{Y_i}$ is a $N
\times |Y_i|$ matrix containing the columns of the $N\times N$
identity corresponding to items in $Y_i$; $B_{Y_i}$ simply serves to
map the gradients with respect to $V_{Y_i}$ into the proper positions
in $V$.  This formula allows us to compute the eigenvector derivatives
in time $O(nNk^2)$, where again $k = \max_i |Y_i|$.  (See
\appref{eigvec-updates} for the derivation of \eqref{v-final-em} and
its computational complexity.)

\eqref{v-final-em} is only valid for the first gradient step, so in practice
we do not bother to fully optimize $V$ in each M-step; we simply take
a single gradient step on $V$.  Ideally we would repeatedly evaluate
the M-step objective, \eqref{mstep-objective}, with various step sizes
to find the optimal one.  However, the M-step objective is intractable
to evaluate exactly, as it is an expectation with respect to an
exponential-size distribution.  In practice, we solve this issue by
performing an E-step for each trial step size.  That is, we update
$q$'s distribution to match the updated $V$ and $\Lambda$ that define
$p_K$, and then determine if the current step size is good by checking
for improvement in the likelihood $\L$.

There is also the issue of enforcing the non-convex constraint
$V^{\top}V = I$.  We could project $V$ to ensure this constraint, but,
as previously discussed for eigen\emph{values}, projection steps often
lead to poor local optima.  Thankfully, for the particular constraint
associated with $V$, more sophisticated update techniques exist: the
constraint $V^{\top}V = I$ corresponds to optimization over a Stiefel
manifold, so the algorithm from \cite[Page 326]{edelman1998sjmaa} can
be employed.  In practice, we simplify this algorithm by negelecting
second-order information (the Hessian) and using the fact that the $V$
in our application is full-rank.  With these simplifications, the
following multiplicative update is all that is needed:
\begin{equation}
V \leftarrow V \exp\left[\eta \left(V^{\top}\df{\L}{V} - \left(\df{\L}{V}\right)^{\top} V\right)\right]~,
\eqlabel{stiefel}
\end{equation}
where $\exp$ denotes the matrix exponential and $\eta$ is the step
size.  \algref{em} summarizes the overall EM method.  As previously
mentioned, assuming $T_1$ iterations until convergence and an average
of $T_2$ iterations to find a step size, its overall runtime is
$O(T_1nNk^2 + T_1T_2N^3)$.  The first term in this complexity comes
from the eigenvalue updates, \eqref{s-vals}, and the eigenvector
derivative computation, \eqref{v-final-em}.  The second term comes
from repeatedly computing the Stiefel manifold update of
$V$, \eqref{stiefel}, during the step size search.

\section{Experiments}
\seclabel{experiments}

We test the proposed EM learning method (\algref{em}) by comparing it
to K-Ascent (\KA{}, \algref{ka})\footnote{Code and data for all
experiments can be downloaded
from \href{https://code.google.com/p/em-for-dpps/wiki/README}
{https://code.google.com/p/em-for-dpps}}.  Both methods require a
starting marginal kernel $\tilde{K}$.  Note that neither \EM{}
nor \KA{} can deal well with starting from a kernel with too many
zeros.  For example, starting from a diagonal kernel, both
gradients, \eqsref{ka-gradient}{v-final-em}, will be diagonal,
resulting in no modeling of diversity.  Thus, the two initialization
options that we explore have non-trivial off-diagonals.  The first of
these options is relatively naive, while the other incorporates
statistics from the data.

For the first initialization type, we use a Wishart distribution with
$N$ degrees of freedom and an identity covariance matrix to draw
$\tilde{L} \sim \W_N(N, I)$.  The Wishart distribution is relatively
unassuming: in terms of eigenvectors, it spreads its mass uniformly
over all unitary matrices \cite{james1964ams}.  We make just one
simple modification to its output to make it a better fit for
practical data: we re-scale the resulting matrix by $1/N$ so that the
corresponding DPP will place a non-trivial amount of probability mass
on small sets.  (The Wishart's mean is $NI$, so it tends to
over-emphasize larger sets unless we re-scale.)  We then convert
$\tilde{L}$ to $\tilde{K}$ via \eqref{k-l-relation}.

For the second initialization type, we employ a form of moment
matching.  Let $m_i$ and $m_{ij}$ represent the normalized frequencies
of single items and pairs of items in the training data:
\begin{equation}
m_i = \frac{1}{n} \sum_{\ell = 1}^n \ind(i \in Y_{\ell}),\;\;\;\;
m_{ij} = \frac{1}{n} \sum_{\ell = 1}^n \ind(i \in Y_{\ell} \wedge j \in Y_{\ell})\,.
\eqlabel{data-marginals}
\end{equation}
Recalling \eqref{k-marginal}, we attempt to match the first and second
order moments by choosing $\tilde K$ as:
\begin{equation}
\tilde{K}_{ii} = m_i,\;\;\;\;
\tilde{K}_{ij} = \sqrt{\max\left(\tilde{K}_{ii}\tilde{K}_{jj} - m_{ij}, 0\right)}\,.
\eqlabel{mm-init}
\end{equation}
To ensure a valid starting kernel, we then project $\tilde{K}$ by
clipping its eigenvalues at $0$ and $1$.

\subsection{Baby registry tests}
\seclabel{babies}

Consider a product recommendation task, where the ground set comprises
$N$ products that can be added to a particular category (e.g., toys or
safety) in a baby registry.  A very simple recommendation system might
suggest products that are popular with other consumers; however, this
does not account for negative interactions: if a consumer has already
chosen a carseat, they most likely will not choose an additional
carseat, no matter how popular it is with other consumers. DPPs are
ideal for capturing such negative interactions.  A learned DPP could
be used to populate an initial, basic registry, as well as to provide
live updates of product recommendations as a consumer builds their
registry.

To test our DPP learning algorithms, we collected a dataset consisting
of $29{,}632$ baby registries from Amazon.com, filtering out those
listing fewer than $5$ or more than $100$ products.  Amazon
characterizes each product in a baby registry as belonging to one of
$18$ categories, such as ``toys'' and``safety''.  For each registry,
we created sub-registries by splitting it according to these
categories.  (A registry with $5$ toy items and $10$ safety items
produces two sub-registries.)  For each category, we then filtered
down to its top $100$ most frequent items, and removed any product
that did not occur in at least $100$ sub-registries.  We discarded
categories with $N < 25$ or fewer than $2N$ remaining (non-empty)
sub-registries for training.  The resulting $13$ categories have an
average inventory size of $N = 71$ products and an average number of
sub-registries $n = 8{,}585$.  We used $70$\% of the data for training
and $30$\% for testing.  Note that categories such as ``carseats''
contain more diverse items than just their namesake; for instance,
``carseats'' also contains items such as seat back kick protectors and
rear-facing baby view mirrors.  See \appref{baby-details} for more
dataset details and for quartile numbers for all of the experiments.

\begin{figure}
  \centering
\subfloat[][]{
  \includegraphics[width=0.47\linewidth,page=1,trim=90 120 70 150,clip]{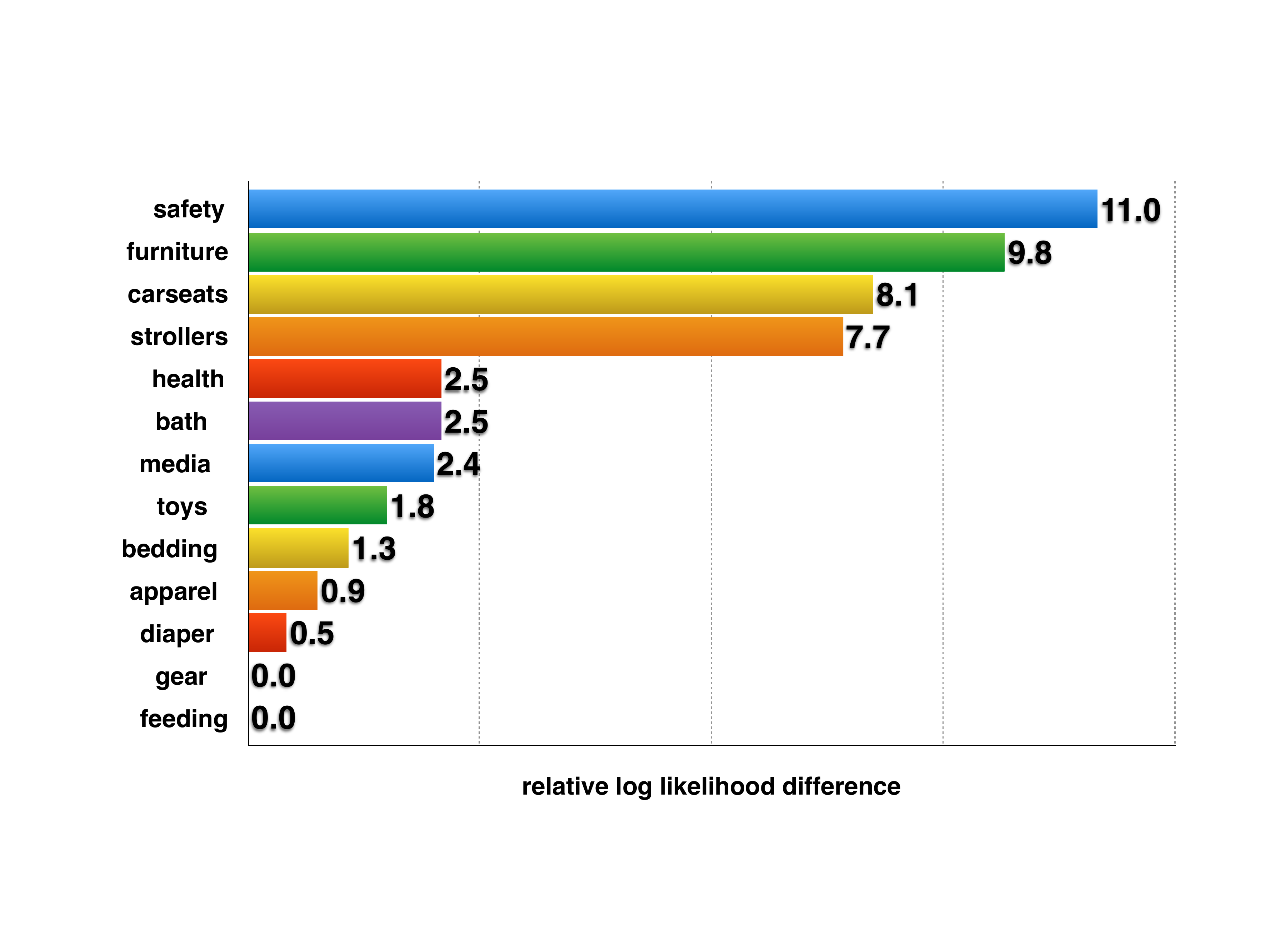}
   \figlabel{regs-wi}
}
\hspace{+0.1in}
\subfloat[][]{
 \includegraphics[width=0.47\linewidth,page=2,trim=90 120 70 150,clip]{plots.pdf}
 \figlabel{regs-mm}
}
\caption{Relative test log-likelihood differences, $100 \frac{(\textrm{EM} -
    \textrm{KA})}{|\textrm{KA}|}$, using: (a) Wishart initialization
    in the full-data setting, and (b) moments-matching initialization
    in the data-poor setting.}
\end{figure}

\figref{regs-wi} shows the relative test log-likelihood differences of
EM and KA when starting from a Wishart initialization.  These numbers
are the medians from $25$ trials (draws from the Wishart).  EM gains
an average of $3.7$\%, but has a much greater advantage for some
categories than for others.  Speculating that EM has more of an
advantage when the off-diagonal components of $K$ are truly
important---when products exhibit strong negative interactions---we
created a matrix $M$ for each category with the true data marginals
from
\eqref{data-marginals} as its entries.  We then checked the value of
$d = \frac{1}{N}\frac{||M||_F}{||\diag(M)||_2}$.  This value
correlates well with the relative gains for EM: the $4$ categories for
which EM has the largest gains (safety, furniture, carseats, and
strollers) all exhibit $d > 0.025$, while categories such as feeding
and gear have $d < 0.012$.  Investigating further, we found that, as
foreshadowed in \secref{baseline}, KA performs particularly poorly in
the high-$d$ setting because of its projection step---projection can
result in KA learning a near-diagonal matrix.

If instead of the Wishart initialization we use the moments-matching
initializer, this alleviates KA's projection problem, as it provides a
starting point closer to the true kernel.  With this initializer, KA
and EM have comparable test log-likelihoods (average EM gain of
0.4\%).  However, the moments-matching initializer is not a perfect
fix for the KA algorithm in all settings.  For instance, consider a
data-poor setting, where for each category we have only $n = 2N$
training examples.  In this case, even with the moments-matching
initializer EM has a significant edge over KA, as shown
in \figref{regs-mm}: EM gains an average of $4.5$\%, with a maximum
gain of $16.5$\% for the safety category.

To give a concrete example of the advantages of EM
training, \figref{safety-sets} shows a greedy
approximation \cite[Section 4]{nemhauser1978mp} to the most-likely
ten-item registry in the category ``safety'', according to a
Wishart-initialized EM model.  The corresponding KA selection differs
from \figref{safety-sets} in that it replaces the lens filters and the
head support with two additional baby monitors: ``Motorola MBP36
Remote Wireless Video Baby Monitor'', and ``Summer Infant Baby Touch
Digital Color Video Monitor''.  It seems unlikely that many consumers
would select three different brands of video monitor.

\begin{figure}
\centering
\subfloat[][]{
\includegraphics[width=0.6\linewidth]{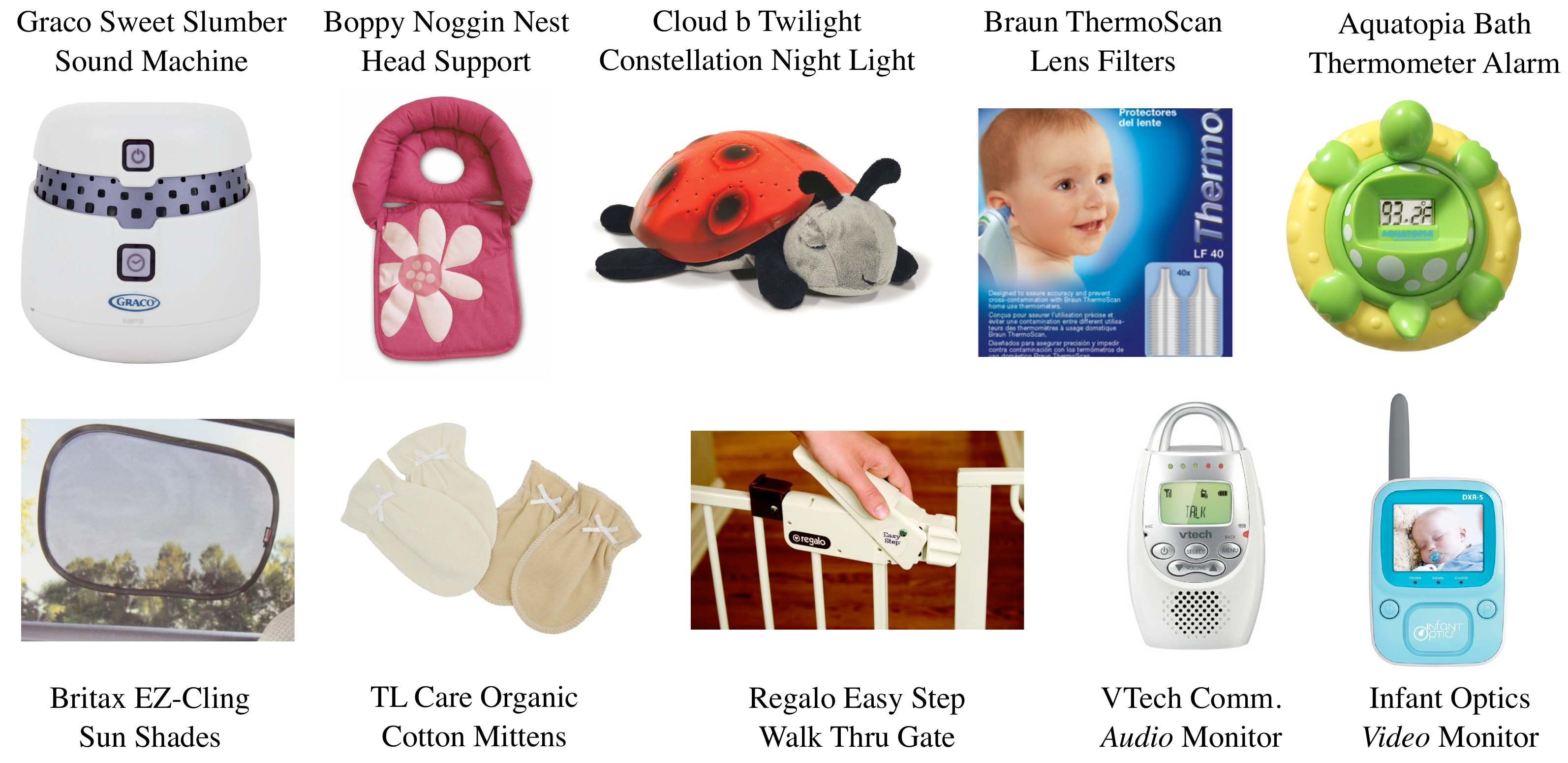}
\figlabel{safety-sets}
}
\subfloat[][]{
 \includegraphics[width=0.4\linewidth,page=3,trim=160 120 150 150,clip]{plots.pdf}
 \figlabel{runtimes}
}
\caption{(a) A high-probability set of size $k = 10$ selected using an EM model for the ``safety'' category.  (b) Runtime ratios.}
\end{figure}

Having established that EM is more robust than KA, we conclude with an
analysis of runtimes.  \figref{runtimes} shows the ratio of KA's
runtime to EM's for each category.  As discussed earlier, EM is
asymptotically faster than KA, and we see this borne out in practice
even for the moderate values of $N$ and $n$ that occur in our
registries dataset: on average, EM is $2.1$ times faster than KA.

\section{Conclusion}
\seclabel{conclusion}

We have explored learning DPPs in a setting where the kernel $K$ is
not assumed to have fixed values or a restrictive parametric form.  By
exploiting $K$'s eigendecomposition, we were able to develop a novel
EM learning algorithm.  On a product recommendation task, we have
shown EM to be faster and more robust than the naive approach of
maximizing likelihood by projected gradient.  In other applications
for which modeling negative interactions between items is important,
we anticipate that EM will similarly have a significant advantage.

\subsubsection*{Acknowledgments}

This work was supported in part by ONR Grant N00014-10-1-0746.

\appendix

\section{Proof of Lemma 1}
\applabel{lemma1-proof}

\lemref{q-is-kdpp} gives the exact form of $q$'s kernel.  Before
giving the proof, we briefly note that $q$ differs slightly from the
typical DPPs we have seen thus far, in its \emph{conditional} nature.
More precisely, for a set $Y_i$ of size $k$, $q$ qualifies as a
\emph{$k$-DPP}, a DPP conditioned on sets of size $k$.  Formally, a
$k$-DPP with (non-marginal) kernel $L$ assigns probability $\propto
\det(L_Y)$ for $|Y| = k$, and probability zero for $|Y| \neq k$.  As
for regular DPPs, a $k$-DPP can be efficiently sampled from and
marginalized, via modifications of the standard DPP algorithms.  For
example, the normalization constant for a $k$-DPP is given by the
identity $\sum_{Y : |Y| = k} \det(L_{Y}) = e_k^N(L)$, where $e_k^N(L)$
represents the $k$th-order elementary symmetric polynomial on the
eigenvalues of $L$ \cite{kulesza2011icml}.  \cite{baker1996apm}'s
``summation algorithm'' computes $e_k^N(L)$ in $O(Nk)$ time.  In short
$k$-DPPs enjoy many of the advantages of DPPs.  Their identical
parameterization, in terms of a single kernel, makes our E-step
simple, and their normalization and marginalization properties are
useful for the M-step updates.

\begin{proof}
Since the E-step is an unconstrained KL divergence minimization, we have:
\begin{equation}
q(J \mid Y_i) = p_K(J \mid Y_i) = \frac{p_K(J, Y_i)}{p_K(Y_i)}
 \propto p_K(J, Y_i) = p_K(J) p_K(Y_i \mid J)
\eqlabel{q-prop}
\end{equation}
where the proportionality follows because $Y_i$ is held constant in
the conditional $q$ distribution.  Recalling \eqref{v-representation},
notice that $p_K(Y_i \mid J)$ can be re-expressed as follows:
\begin{equation}
  p_K(Y_i \mid J) = 
\ind(|Y_i| = |J|) \det([V^J (V^J)^\top]_{Y_i}) = 
\ind(|Y_i| = |J|) \det([U^{Y_i} (U^{Y_i})^\top]_J)\,.
  \eqlabel{u-representation}
\end{equation}
This follows from the identity $\det(AA^{\top}) = \det(A^{\top}A)$, for
any full-rank square matrix $A$.  The subsequent swapping of $J$ and
$Y_i$, once $V^{\top}$ is re-written as $U$, does not change the
indexed submatrix.

Plugging this back into \eqref{q-prop}:
\begin{equation}
q(J \mid Y_i) \propto p_K(J) \ind(|Y_i| = |J|) \det([U^{Y_i} (U^{Y_i})^\top]_J)
= p_K(J) \ind(|Y_i| = |J|) P^{U^{Y_i}}(J) \\
\end{equation}
where $P^{U^{Y_i}}$ represents an elementary DPP, just as in
\eqref{elem-kern}, but over $J$ rather than $Y$.  Multiplying this
expression by a term that is constant for all $J$ maintains
proportionality and allows us to simplify the the $p_K(J)$ term.
Taking the definition of $p_K(J)$ from \eqref{v-representation}:
\begin{align}
q(J \mid Y_i) \propto&\; \left(\prod_{j = 1}^N \frac{1}{1 - \lambda_j}\right) \ind(|Y_i| = |J|) P^{U^{Y_i}}(J) \prod_{j \in J} \lambda_j \prod_{j \notin J} (1 - \lambda_j) \\
=&\; \ind(|Y_i| = |J|) P^{U^{Y_i}}(J) \prod_{j \in J} \frac{\lambda_j}{1 - \lambda_j}
\end{align}
Having eliminated all dependence on $j \notin J$, it is now possible
to express $q(J \mid Y_i)$ as the $J$ principal minor of a PSD kernel
matrix (see $Q^{Y_i}$ in the statement of the lemma).  Thus, $q$
is a $k$-DPP.
\end{proof}

\newpage

\section{M-Step eigenvalue updates}
\applabel{eigval-updates}

We can exploit standard $k$-DPP marginalization formulas to
efficiently compute the eigenvalue updates for EM.  Specifically, the
exponential-size sum over $J$ from \eqref{lam-Jexp-up} can be reduced
to the computation of an eigendecomposition and several elementary
symmetric polynomials on the resulting eigenvalues.  Let $e_{k -
  1}^{-j}(Q^{Y_i})$ be the $(k - 1)$-order elementary symmetric
polynomial over all eigenvalues of $Q^{Y_i}$ except for the $j$th one.
Then, by direct application of \cite[Equation 205]{kulesza2012ftml},
$q$'s singleton marginals are:
\begin{equation}
\sum_{J : j \in J} q(J \mid Y_i) =
q(j \in J \mid Y_i) = \frac{1}{e_{|Y_i|}^N(Q^{Y_i})} \sum_{r = 1}^N
  \hat{v}_r(j)^2 \hat{\lambda}_r e_{|Y_i| - 1}^{-r}(Q^{Y_i})\,.
\eqlabel{ugly-lam-up}
\end{equation}
As previously noted, elementary symmetric polynomials can be
efficiently computed using \cite{baker1996apm}'s ``summation
algorithm''.

We can further reduce the complexity of this formula by noting that
rank of the $N \times N$ matrix $Q^{Y_i} = RZ^{Y_i}R$ is at most
$|Y_i|$.  Because $Q^{Y_i}$ only has $|Y_i|$
 non-zero eigenvalues, it is the case that, for all $r$:
\begin{equation}
\hat{\lambda}_r e_{|Y_i| - 1}^{-r}(Q^{Y_i}) = e_{|Y_i|}^N(Q^{Y_i})\,.
\end{equation}
Recalling that the eigenvectors and eigenvalues of $Q^{Y_i}$ are
denoted $\hat{V}, \hat{\Lambda}$, the computation of the singleton
marginals of $q$ that are necessary for the M-step eigenvalue updates
can be written as follows:
\begin{equation}
q(j \in J \mid Y_i) = \frac{1}{e_{|Y_i|}^N(Q^{Y_i})} \sum_{r = 1}^N 
  \hat{v}_r(j)^2 \hat{\lambda}_r e_{|Y_i| - 1}^{-r}(Q^{Y_i}) = \sum_{r = 1}^{|Y_i|} \hat{v}_r(j)^2\,.
\eqlabel{eigvals-via-H}
\end{equation}

This simplified formula is dominated by the $O(N^3)$ cost of the
eigendecompositon required to find $\hat{V}$.  This cost can be
further reduced, to $O(Nk^2)$, by eigendecomposing a related matrix
instead of $Q^{Y_i}$.  Specifically, consider the $|Y_i| \times |Y_i|$
matrix $H^{Y_i} = V_{Y_i} R^2 V_{Y_i}^{\top}$.  Let $\tilde{V}$ and
$\tilde{\Lambda}$ be the eigenvectors and eigenvalues of $H^{Y_i}$.
This $\tilde{\Lambda}$ is identical to the non-zero eigenvalues of
$Q^{Y_i}$, $\hat{\Lambda}$, and its eigenvectors are related as
follows:
\begin{equation}
\hat{V} = R V_{Y_i}^{\top} \tilde{V} \diag\left(\frac{\bone}{\sqrt{\tilde{\blambda}}}\right)\,.
\eqlabel{hV-via-tV}
\end{equation}
Getting $\hat{V}$ via \eqref{hV-via-tV} is an $O(N|Y_i|^2)$ operation,
given the eigendecomposition of $H^{Y_i}$.  Since this
eigendecomposition is an $O(|Y_i|^3)$ operation, it is dominated by
the $O(N|Y_i|^2)$.  To compute \eqref{eigvals-via-H} for all $j$ and
requires only $O(Nk)$ time, given $\hat{V}$.  Thus, letting $k =
\max_{i} |Y_i|$, the size of the largest example set, the overall
complexity of the eigenvalue updates is $O(nNk^2)$.

\newpage

\section{M-Step eigenvector gradient}
\applabel{eigvec-updates}

Recall that the M-step objective is:
\begin{equation}
F(V, \Lambda) = \sum_{i = 1}^n \sum_{J} q(J \mid Y_i) \left[
\log p_K(J) + \log p_K(Y_i \mid J) \right]\,.
\end{equation}
The $p_K(J)$ term does not depend on the eigenvectors, so we only have
to be concerned with the $p_K(Y_i \mid J)$ term when computing the
eigenvector derivatives.  Recall that this term is defined as follows:
\begin{equation}
p_K(Y_i \mid J) = \ind(|Y_i| = |J|) \det\left(\left[V^J (V^J)^\top\right]_{Y_i}\right)\,.
\eqlabel{y-given-j}
\end{equation}
Applying standard matrix derivative rules such as \cite[Equation
  55]{matrixcookbook}, the gradient of the M-step objective with
respect to entry $(a,b)$ of $V$ is:
\begin{equation}
\df{F(V, \Lambda)}{[V]_{ab}} = \sum_{i = 1}^n 
\sum_J q(J \mid Y_i) \ind(a \in Y_i \wedge b \in J) 
2 [(W^{J}_{Y_i})^{-1}]_{g_{Y_i}(a)} \cdot \v_{b}(Y_i)
\end{equation}
where $W^J_{Y_i} = [V^{J} (V^{J})^T]_{Y_i}$ and the subscript
$g_{Y_i}(a)$ indicates the index of $a$ in $Y_i$.  The
$[(W^{J}_{Y_i})^{-1}]_{g_{Y_i}(a)}$ indicates the corresponding row in
$W^{J}_{Y_i}$, and $\v_b(Y_i)$ is eigenvector $b$ restricted to
$Y_i$.  Based on this, we can more simply express the derivative with
respect to the entire $V$ matrix:
\begin{equation}
\df{F(V, \Lambda)}{V} = \sum_{i = 1}^n \sum_J 2 q(J \mid Y_i)
(\dot{W}^J_{Y_i})^{-1} \dot{V}
\end{equation}
where the $\dot{V} = \diag(\bone_{Y_i}) V \diag(\bone_{J})$ is equal
to $V$ with the rows $\ell \notin Y$ and the columns $j \notin J$
zeroed.  Similarly, the other half of the expression represents
$(W^J_{Y_i})^{-1}$ sorted such that $g_{Y_i}(\ell) = \ell$ and
expanded with zero rows for all $\ell \notin Y_i$ and zero columns for
all $\ell \notin Y_i$.  The exponential-size sum over $J$ could be
approximated by drawing a sufficient number of samples from $q$, but
in practice that proves somewhat slow.  It turns out that it is
possible, by exploiting the relationship between $Z$ and $V$, to
perform the first gradient step on $V$ without needing to sample $q$.

\subsection{Exact computation of the first gradient}

Recall that $Z^{Y_i}$ is defined to be $U^{Y_i} (U^{Y_i})^\top$, where
$U = V^{\top}$.  The $p_K(Y_i \mid J)$ portion of the M-step
objective, \eqref{y-given-j}, can be re-written in terms of $Z^{Y_i}$:
\begin{equation}
p_K(Y_i \mid J) = \ind(|Y_i| = |J|) \det\left(Z^{Y_i}_J\right)\,.
\end{equation}
Taking the gradient of the M-step objective with respect to $Z^{Y_i}$:
\begin{equation}
\df{F(V, \Lambda)}{Z^{Y_i}} = \sum_J q(J \mid Y_i) (Z^{Y_i}_J)^{-1}\,.
\end{equation}
Plugging in the $k$-DPP form of $q(J \mid Y_i)$ derived in the main
body of the paper:
 \begin{equation}
\df{F(V, \Lambda)}{Z^{Y_i}} =
\frac{1}{e_{|Y_i|}^N(Q^{Y_i})} 
\sum_{J : |J| = |Y_i|} \det(Q^{Y_i}_J) (Z^{Y_i}_J)^{-1}\,.
\eqlabel{m-grad-adj}
\end{equation}
Recall from the background section the identity used to normalize a
$k$-DPP, and consider taking its derivative with respect to $Z^{Y_i}$:
\begin{equation}
\sum_{J : |J| = k} \det(Q^{Y_i}_J) = e_k^N(Q^{Y_i})
\quad \underset{\textrm{derivative wrt $Z^{Y_i}$}}\Longrightarrow \quad
\sum_{J : |J| = k} \det(Q^{Y_i}_J) (Z^{Y_i}_J)^{-1} = \df{e_k^N(Q^{Y_i})}{Z^{Y_i}}
\eqlabel{esp-adj-ident}
\end{equation}
Note that this relationship is only true at the start of the M-step,
before $V$ (and hence $Z$) undergoes any gradient updates; a gradient
step for $V$ would mean that $Q^{Y_i}$, which remains fixed during the
M-step, could no longer can be expressed as $RZ^{Y_i}R$.  Thus, the
formula we develop in this section is only valid for the first
gradient step.

Plugging \eqref{esp-adj-ident} back into \eqref{m-grad-adj}:
\begin{equation}
\df{F(V, \Lambda)}{Z^{Y_i}} =
\frac{1}{e_{|Y_i|}^N(Q^{Y_i})} 
\frac{\partial e_{|Y_i|}^N(Q^{Y_i})}{\partial Z^{Y_i}}\,.
\eqlabel{esp-v-grad}
\end{equation}

Multiplying this by the derivative of $Z^{Y_i}$ with respect to $V$
and summing over $i$ gives the final form of the gradient with respect
to $V$.  Thus, we can compute the value of the first gradient on $V$
exactly in polynomial time.

\subsection{Faster computation of the first gradient}

Recall from \appref{eigval-updates} that the rank of the $N \times N$
matrix $Q^{Y_i} = RZ^{Y_i}R$ is at most $|Y_i|$ and that its non-zero
eigenvalues are identical to those of the $|Y_i| \times |Y_i|$ matrix
$H^{Y_i} = V_{Y_i} R^2 V_{Y_i}^{\top}$.  Since the elementary
symmetric polynomial $e_k^N$ depends only on the eigenvalues of its
argument, this means $H^{Y_i}$ can substitute for $Q^{Y_i}$ in
\eqref{esp-v-grad}, if we change variables back from $Z$ to $V$:
\begin{equation}
\df{F(V, \Lambda)}{V} = \sum_{i = 1}^n
\frac{1}{e_{|Y_i|}^{N}(H^{Y_i})} 
\frac{\partial e_{|Y_i|}^{N}(H^{Y_i})}{\partial V}
\eqlabel{h-it-all}
\end{equation}
where the $i$-th term in the sum is assumed to index into the $Y_i$
rows of the $V$ derivative.  Further, because $H$ is only size $|Y_i|
\times |Y_i|$:
\begin{equation}
e_{|Y_i|}^{N}(H^{Y_i}) = e_{|Y_i|}^{|Y_i|}(H^{Y_i}) = \det(H^{Y_i})\,.
\end{equation}
Plugging this back into \eqref{h-it-all} and applying standard matrix
derivative rules:
\begin{equation}
\df{F(V, \Lambda)}{V} = \sum_{i = 1}^n
\frac{1}{\det(H^{Y_i})} 
\frac{\partial \det(H^{Y_i})}{\partial V} = 
 \sum_{i = 1}^n 2 (H^{Y_i})^{-1} V_{Y_i} R^2\,.
\eqlabel{efficient-V-grad}
\end{equation}
Thus, the initial M-step derivative with respect to $V$ can be more
efficiently computed via the above equation.  Specifically, the matrix
$H^{Y_i}$ can be computed in time $O(N|Y_i|^2)$, since $R$ is a
diagonal matrix.  It can be inverted in time $O(|Y_i|^3)$, which is
dominated by $O(N|Y_i|^2)$.  Thus, letting $k = \max_{i} |Y_i|$, the
size of the largest example set, the overall complexity of computing
the eigenvector gradient in \eqref{efficient-V-grad} is $O(nNk^2)$.

\newpage

\section{Baby registry details}
\applabel{baby-details}

\figref{reg-data} and \figref{reg-quantiles} contain details, referred
to in the main body of the paper paper, about the baby registry
dataset and the learning methods' performance on it.

\begin{figure}[h]
  \centering
\subfloat[][]{
\begin{tabular}{|l|r|r|} \hline
\bf Category  & \bf N    & \bf \# of Regs \\ \hline
feeding   & 100  & 13300  \\ 
gear      & 100  & 11776 \\
diaper    & 100  & 11731  \\
bedding   & 100  & 11459  \\
apparel   & 100  & 10479  \\
bath      & 100  & 10179  \\
toys      & 62   & 7051   \\
health    & 62   & 9839   \\
media     & 58   & 4132   \\
strollers & 40   & 5175   \\
safety    & 36   & 6224   \\
carseats  & 34   & 5296   \\
furniture & 32   & 4965   \\ \hline
\end{tabular}
\figlabel{reg-data}
}
\hspace{0.1in}
\subfloat[][]{
\begin{tabular}{|l|ccc|c|ccc|} \hline
\multirow{2}{*}{\bf Category}  & \multicolumn{3}{|c|}{\bf Wishart} & \multicolumn{1}{|c|}{\bf Moments} & \multicolumn{3}{|c|}{\bf Moments} \\
              &           &           &            & (all data) &            & (less data) & \\ \hline
safety        &  (10.88)  &   11.05   &   (11.12)  &  -0.13     &  (10.19)   &  16.53      &  (19.46)    \\ 
furniture     &   (9.80)  &    9.89   &   (10.07)  &   0.23     &   (8.00)   &  10.47      &  (13.57)    \\
carseats      &   (8.06)  &    8.16   &   (8.31)   &   0.61     &   (3.40)   &   5.85      &   (8.28)    \\
strollers     &   (7.66)  &    7.77   &   (7.88)   &  -0.07     &   (2.51)   &   5.35      &   (7.41)    \\
health        &   (2.50)  &    2.54   &   (2.58)   &   1.37     &   (2.67)   &   5.36      &   (6.03)    \\
bath          &   (2.50)  &    2.54   &   (2.59)   &  -0.24     &   (2.22)   &   3.56      &   (4.23)    \\
media         &   (2.37)  &    2.42   &   (2.49)   &  -0.17     &   (0.44)   &   1.93      &   (2.77)    \\
toys          &   (1.76)  &    1.80   &   (1.83)   &   0.13     &   (1.01)   &   2.39      &   (4.30)    \\
bedding       &   (0.42)  &    1.34   &   (1.44)   &   2.81     &   (2.44)   &   3.19      &   (3.70)    \\
apparel       &   (0.88)  &    0.92   &   (0.93)   &   0.53     &   (0.78)   &   1.59      &   (2.23)    \\
diaper        &   (0.50)  &    0.58   &   (1.02)   &  -0.47     &  (-0.87)   &  -0.19      &   (1.26)    \\
gear          &   (0.03)  &    0.05   &   (0.07)   &   0.86     &   (1.36)   &   2.63      &   (3.22)    \\
feeding       &  (-0.11)  &   -0.09   &  (-0.07)   &  -0.03     &  (-1.32)   &   0.61      &   (1.22)    \\ \hline
average       &           &    3.76   &            &   0.41     &            &   4.55      &              \\ \hline
\end{tabular}
\figlabel{reg-quantiles}
}
\caption{(a) Size of the post-filtering ground set for each product
  category, and the associated number of sub-registries (subsets of
  $\{1, \ldots, N\}$).  (b) Relative test log-likelihood differences,
  $100\frac{(\textrm{EM} - \textrm{KA})}{|\textrm{KA}|}$, for three
  cases: a Wishart initialization, a moments-matching initialization,
  and a moments-matching initialization in a low-data setting (only $n
  = 2N$ examples in the training set).  For the first and third
  settings there is some variability: in the first setting, because
  the starting matrix drawn from the Wishart can vary; in the third
  setting, because the training examples (drawn from the full training
  set used in the other two settings) can vary.  Thus, for these two
  settings the numbers in parentheses give the first and third
  quartiles over 25 trials.}
\end{figure}

\newpage

\bibliographystyle{unsrt}
\bibliography{paper}

\end{document}